\documentclass[review]{elsarticle}

\usepackage{hyperref}
\usepackage{amssymb,amsfonts}
\usepackage{amsmath}
\usepackage{algorithm}
\usepackage{algorithmic}
\usepackage{graphicx}
\usepackage{textcomp}
\usepackage{subfigure}
\usepackage{xcolor}
\usepackage{setspace}
\usepackage{ulem}
\usepackage{multirow}
\usepackage{tabularx}
\usepackage{multirow}
\usepackage{url}
\journal{Expert systems with applications}

\biboptions{authoryear}

\begin{document}

\begin{frontmatter}

\title{\textit{initKmix}-A Novel Initial Partition Generation Algorithm for Clustering Mixed Data using k-means-based Clustering}


\author{Amir Ahmad}\corref{mycorrespondingauthor}
\address{College of Information Technology,
              United Arab Emirates University,Al-Ain, UAE
              }
\cortext[mycorrespondingauthor]{Corresponding author}
\ead{amirahmad@uaeu.ac.ae}
             
\author{Shehroz Khan}
\address{KITE -- Toronto Rehabilitation Institute, 
              University Health Network,
              550, University Avenue, 
              Toronto, ON, Canada}
\ead{shehroz.khan@uhn.ca}

\begin{abstract}
Mixed datasets consist of both numeric and categorical attributes. Various k-means-based clustering algorithms have been developed for these datasets. Generally, these  algorithms use random partition  as a starting point, which tends to produce different clustering results for different runs.  In this paper, we propose, \textit{initKmix}, a novel algorithm for finding an initial partition for k-means-based clustering algorithms for mixed datasets. In the \textit{initKmix} algorithm, a k-means-based clustering algorithm is run many times, and in each run, one of the  attributes is used to create initial clusters for that run. The clustering results of various runs are combined to produce the initial partition. This initial partition is then used as a seed to a k-means-based clustering algorithm to cluster mixed data. Experiments with various categorical and mixed datasets showed that  \textit{initKmix} produced accurate and consistent results, and outperformed  the random initial partition method and other state-of-the-art initialization methods. Experiments also showed that k-means-based clustering for mixed datasets with \textit{initKmix} performed similar to or better than many state-of-the-art clustering algorithms for categorical and mixed datasets.
\end{abstract}

\begin{keyword}
\texttt{Mixed data}\sep Clustering \sep K-means \sep Initialization \sep Random
\MSC[2010] 00-01\sep  99-00
\end{keyword}

\end{frontmatter}

%\linenumbers
\section {Introduction} 
\label{sec:intro}
Clustering is a process in which similar data points are grouped in the same clusters whereas dissimilar data points are grouped is different clusters based on some notion of `similarity' \citep{ClusteringbookJain}. %In clustering, $n$ data points of a dataset having $m$ attributes are grouped in $K$ clusters. 
Many datasets only contain numeric or categorical attributes; however, the majority of real-world datasets contain both types of attributes. These are called mixed datasets \citep{Booksmachine,WEKAwitten1}. Most clustering algorithms developed to handle only numeric or categorical datasets cannot be directly used to cluster mixed datasets because the calculation of `similarity' is not straight forward in mixed datasets \citep{amirshehrozsurvey}.

Various types of clustering algorithms have been developed to handle mixed datasets \citep{mixedddatareview}; the most prominent being the partitional, hierarchical, model-based, and neural network-based methods \citep{amirshehrozsurvey}. The partitional clustering methods are more popular among the research community because they are (i)
simpler in interpretation and implementation, (ii) linear in time complexity with the number of data points, and (iii) easily adaptable to parallel architectures. The traditional k-means algorithm is a partitional clustering algorithm that was developed to cluster datasets with only numeric attributes \citep{kmeanclusterinitialization1}. Its objective function uses a distance metric (e.g. Euclidean distance) that can only be defined for numeric attributes. This type of clustering algorithm is further extended to develop  \textbf{k}-means-based clustering  for \textbf{m}ixed \textbf{d}atasets (KMD) algorithms \citep{Huang1997,AmirLipika2007,amirshehrozsurvey}. The KMD algorithms comprise of a family of algorithms that may differ in the definition of their cluster center, distance measure and objective function \citep{amirshehrozsurvey}. \textcolor{black}{Most of these partitional clustering approaches perform hard clustering, i.e. a data point can belong to only one cluster. In fuzzy clustering approaches, a data point can be assigned to more than one cluster with different membership values. Approaches based on fuzzy clustering have also been applied for mixed datasets \citep{ji2012fuzzy,DURSO2019513}. In this paper, we focus on hard clustering clustering algorithms.}% that assign a data point to only one cluster.}  
%However, it has been modified to handle 
%K-means-based clustering algorithms are popular clustering approach to handle these kinds of datasets  because of low computational complexity \cite{Huang1997,AmirLipika2007,amirshehrozsurvey}. 

In general, there are two type of approaches for initializing the k-means clustering algorithm. In the first approach, the random initial cluster centers,  \textit{k} (the number of clusters) data points are selected randomly, which act as the initial cluster centers. In the second approach, the random initial partition, first randomly assigns a cluster to each data point and then compute the centers of these clusters. The k-means based clustering algorithms may suffer from several drawbacks \citep{kmeanclusterinitialization1}. Prominently, the k-means optimization function can stuck in the local minima; therefore, with a different initialization, the k-means clustering algorithm  may lead to different final clusters. Hence, it is difficult to obtain reliable and consistent clustering results \citep{kmeanclusterinitialization1}.

 Similar to the standard k-means clustering algorithm for numeric data, KMD algorithms also suffers from the random initialization problem. A few initialization methods have been developed for KMD algorithms  \citep{Clusterinitialization2, Clustercenterinitialization,Jinyin2017539,WANGCHAMHAN,AhmadKharmonic}. \textcolor{black}{ However, these methods are either computationally expensive (quadratic complexity with respect to the number of data points) \citep{Clustercenterinitialization,Clusterinitialization2,Jinyin2017539} or do not produce consistent clustering results \citep{WANGCHAMHAN,AhmadKharmonic}}. In this paper, our baseline KMD algorithm  is the one proposed by Ahmed and Dey (\citeyear{AmirLipika2007}), \textbf{k}-\textbf{m}eans \textbf{c}lustering for \textbf{m}ixed \textbf{d}atasets with a mixed distance measure (KMCMD). The reason to choose KMCMD is that it has shown superior performance in comparison to other similar partitional clustering algorithms to cluster mixed datasets \citep{AmirLipika2007}.

In this paper, we  present the \textit{initKmix} algorithm, a novel algorithm to compute the initial partition for KMCMD algorithm. The initial partition is then fed to KMCMD algorithm to cluster mixed datasets. 
The initial clusters produced by the \textit{initKmix} algorithm are stable across various runs of the algorithm or when the order of the data points is changed. This ensures consistent and reliable clustering results. Moreover, the time complexity of the \textit{initKmix} algorithm is linear w.r.t. the number of data points; thus, it can be used with large datasets. The \textit{initKmix} algorithm does not guarantee to find the global optima of the optimization functions of the KMCMD algorithm; however, the experiments suggest that the KMCMD algorithm with the \textit{initKmix} algorithm produces superior clustering results. The choice of \textit{k}  is an important issue in the KMCMD algorithm \citep{clusternumber}; however, in this paper, we focus on finding the appropriate initial partitions with given value of $k$.

The paper is organized in the following way. Section \ref{sec:related} presents related work focusing on the methods to calculate the initial partition for KMCMD algorithms for mixed datasets. The \textit{initKmix} initial partition method is presented in Section 3. Results are presented in Section 4, followed by conclusions and future research directions in Section 5.

\section{Related Work} 
\label{sec:related}
The k-means clustering algorithm is a commonly used clustering algorithm for datasets consisting of numeric attributes because of its low computational complexity \citep{MacQueen1967}. Its complexity is  linear with respect to the number of data points and scales well for large datasets. The algorithm  minimizes the  following optimization function (Eq. (1)) iteratively,
\begin{eqnarray}
\sum\limits_{i=1}^n\xi (d_i,C_i) 
\end{eqnarray}                    

where $n$ is the number of data points in the dataset,  $C_i$ is the nearest cluster centre to data point $d_i$, $\xi$ is a chosen distance measure (or similarity measure) between $d_i$ and $C_i$. Generally, the Euclidean distance is used as the distance measure.

\begin{algorithm}[!ht]
		\caption{General steps of a KMD algorithm with the random partition.}
	\label{kmeansalgo}
		\begin{algorithmic}
		\STATE {\bf Input-}  Mixed dataset \textit{T},  the number of data points is $n$, the number of attributes is $m$, the number of clusters is $k$.
		\STATE Begin
		\STATE 1- {Assign all data points to $k$ clusters randomly}
		\STATE 2- Repeat steps $A$ and $B$ 
		\STATE (A) Calculate the centres of the the clusters.
		\STATE (B) Each data point is assigned to its nearest cluster using the user defined distance measure.
		\STATE Until data points no longer change cluster membership or a predefined number of iterations is reached.
		\STATE End 
\end{algorithmic}
\end{algorithm} 

k-means clustering algorithm computes cluster centres and data point memberships at each iteration. The algorithm starts with user-defined initial clusters. Generally, a random initial partition is selected that may produce different clustering results for different runs of the algorithm. To overcome this problem various methods have been proposed for the computation of the initial partition \citep{Initilizatiaonduda,Bradley98refininginitial,kmeanclusterinitialization1,Arthur}. 
The k-means clustering algorithm can only handle pure numeric datasets. For pure categorical datasets, k-modes clustering algorithm is proposed \citep{Huang97afast}, in which the cluster centre is represented by the mode of the attribute values of the data points presented in that cluster and the Hamming distance is used to compute the distance between a data point and a cluster centre. Similar to the k-means clustering algorithm, k-modes clustering algorithm can also show inconsistent clustering results due to the choice of  random  initial partition. Various algorithms have been proposed to find the appropriate initial partition for K-modes clustering algorithm. \citep{kModeclusterinitilization1,InitializationCao,InitializationWu,khan2003computing,khan2007computation}.

Several KMD algorithms have been developed to extend the k-means clustering algorithm
to mixed datasets \citep{Huang1997,AmirLipika2007}. In these algorithms, new definitions for cluster centres and distances are proposed to handle mixed datasets \citep{Huang1997,AmirLipika2007}. The general steps in these algorithms algorithm are presented in Algorithm \ref{kmeansalgo}. A comprehensive review of these algorithms can be found in the survey paper of Ahmad and Khan \citep{amirshehrozsurvey}. Due to their similarity with k-means clustering, these algorithms also suffer from the issues of random initialization and finding appropriate number of clusters (k). The focus of this paper is to find initial partition for a KMD algorithm (the KMCMD algorithm \citep{AmirLipika2007}). Therefore, in this section, we restrict our literature survey to the research works that attempted to initialize KMD algorithms.

Ji et al. (\citeyear{Clusterinitialization2}) suggest a method to determine initial clusters for a KMD algorithm by computing compute density and distances between data points. In another work, Ji et al. (\citeyear{Clustercenterinitialization}) propose the concept of centrality based on neighbouring points and combined it with the distances between data points to compute initial clusters. However, these algorithms have quadratic complexity and are therefore may not work efficiently for large mixed datasets.
Using density peaks \citep{Rodriguez1492},
Chen et al. (\citeyear{Jinyin2017539}) propose an algorithm to determine the initial cluster centres for mixed datasets. Higher density points are used
to identify cluster centres. This algorithm also has quadratic complexity.
Wangchamhan et al. (\citeyear{WANGCHAMHAN}) combine a search algorithm, the League Championship algorithm \citep{LeagueChampionshipAlgorithm}, with a KMD algorithm to identify the initial cluster centres. This algorithm has many parameters; hence, the final clustering results  are dependent on parameter settings, where different parameters may lead to different clustering results. 
Zheng et al. (\citeyear{Zheng5586136}) combine the evolutionary algorithm (EA) with the k-prototypes clustering algorithm \citep{Huang1998}. The global searching capability of EA leads the proposed algorithm to be stable to cluster initialization. However, clustering results are dependent upon the parameters and not consistent in different runs.

The k-Harmonic means clustering
algorithm addresses the random initial clusters problem applying
a cost function \citep{ZhangKHarmonic} for numeric datasets which uses
the harmonic means of the distances from each data point to
the centers. This algorithm creates clusters that are  more
stable than those generated by k-means clustering algorithm with random
 initial clusters.
Ahmad and Hashmi (\citeyear{AhmadKharmonic}) combine the distance measure and the definition of cluster centres for mixed datasets  suggested by Ahmad and Dey (\citeyear{AmirLipika2007}) with the k-Harmonic clustering algorithm \citep{ZhangKHarmonic} to develop the k-Harmonic clustering algorithm for mixed datasets. Their method is less sensitive to the choice of the initial cluster centres. The standard deviation of the clustering accuracy of this method is small in comparison to that of the random initialization method.  However,  clustering results are not stable for runs with different initial partitions.

The literature review suggests that the existing initialization methods for KMD algorithms are either computationally expensive or do not produce consistent clustering results. This further limits the use of existing methods in real world situations where an algorithm's execution time and reliability of clustering results are the key factors for their adoption.
In the next section, we present our proposed algorithm  \textit{initKmix} to compute the initial partition for a KMD algorithm (KMCMD algorithm \citep{AmirLipika2007}).

\section{\textcolor{black}{\textit{initKmix}  Algorithm}}
\label{sec:kmcmd}
The \textit{initKmix} algorithm is based on following two experimental observations noted in \textcolor{black}{previous research works in k-means clustering  \citep{kmeanclusterinitialization1} and k-modes clustering \citep{kModeclusterinitilization1}:}
\begin{enumerate}[(i)]
    \item Some data points in a given dataset have similar final cluster membership irrespective of the initial partition \citep{kmeanclusterinitialization1,kModeclusterinitilization1}. This observation has been  used to determine the initial partition for k-means clustering algorithm (for pure numeric datasets) \citep{kmeanclusterinitialization1}  and k-modes clustering algorithm (for pure categorical datasets) \citep{kModeclusterinitilization1}. We extend this approach to determine the initial partition for the KMCMD algorithm.
    \item Each of the attributes of a dataset can contribute to final clustering result. Therefore, an individual attribute may be used to determine the initial partition  \citep{kmeanclusterinitialization1,kModeclusterinitilization1}.
\end{enumerate}

%One of the motivations for this method is that 
\textcolor{black}{The \textit{initKmix} algorithm generates multiple instances of clustering, this aspect is similar to multiple-view clustering \citep{mutiwaymainpaper,multiwayclusteringsurveys}.}
Multiple-view clustering manages the production of different clustering results for datasets generated from different sources or observed from different views. These clustering results are combined to generate a clustering result. The diversity of the various views is an important aspect of this clustering approach. It is suggested that multiple views should be used to present data points more comprehensively and accurately \citep{mutiwaymainpaper,multiwayclusteringsurveys}.  \textit{initKmix} algorithm uses a similar approach as it creates multiple clustering and for each clustering a different view is used to create the initial partition.

\textcolor{black}{The \textit{initKmix} algorithm has two important components, running KMCMD algorithm  $m$ times ($m$ is the number of attributes) to generate $m$ instances of clustering. Each instance of clustering creates a cluster label for each data point. $m$ instances of clustering generate a string of $m$ cluster labels for each data point (Table 1). These $m$ instances of clustering (strings of $m$ cluster labels) are combined to generate a clustering, which is then used as initial partition for the KMCMD algorithm.}
As discussed earlier, a mixed dataset contains two types of attributes; categorical and numeric. We use each of these $m$ attributes to create the initial partition in one of the runs of the KMCMD algorithm. These $m$ runs of the KMCMD algorithm  generate $m$ clustering results. The clustering ensemble algorithm \citep{Strehl} is then used to combine these $m$ results to produce final clustering that is used as the initial partition for the KMCMD algorithm. In one run, one attribute is used to create the initial partition that can be considered as one view of the data. \textcolor{black}{Multiple-views of the data by using different initial partitions are used to create multiple instances of clustering.} These views are diverse as they use different attributes to create the initial partition. Hence, we expect that combining these different clustering results will generate an accurate clustering that can be used as the initial partition for the KMCMD algorithm.
The specific steps are presented in Algorithm \ref{Algorithm}.  We will discuss each step of the proposed method in detail below.

\begin{algorithm}[!ht]
\caption{The proposed \textit{initKmix} algorithm for creating initial partition for the KMCMD algorithm.}
\label{Algorithm}
\begin{algorithmic}
\STATE {\bf Input-}  Mixed dataset {\it T}, the number of attributes is $m$, the number of numeric attributes is $m_r$ and the number of categorical attributes is $m_c$  ($m$ = $m_r$ + $m_c$), $k$ is the number of desired clusters. 
\STATE Begin
\STATE {1(a)- For numeric attributes}
\FOR{i=1...$m_r$} 
\STATE  Create a clustering result by using $KMCMD\_Initial\_Numeric$ algorithm (Algorithm \ref{Algorithmnumeric}). 
\ENDFOR
\STATE {1(b)- For categorical attributes}
\FOR{i=1...$m_c$} 
\STATE Create a clustering result by using $KMCMD\_Initial\_Categorical$ algorithm  (Algorithm \ref{Algorithmcategorical}).
\ENDFOR
\\
/*We will have $m$ clustering results as presented in Table \ref{tabensemb} */
\STATE 2- Combine these $m$ clustering results by using a clustering ensemble algorithm discussed in Section \ref{sec:combining}  to get $k$ clusters. 
\\
/*These $k$ clusters will be treated as the initial partition for the KMCMD algorithm.*/  
\STATE End         
\end{algorithmic}
\end{algorithm}

\begin{algorithm}[!ht]
\caption{ $KMCMD\_Initial\_Numeric$ algorithm for creating clusters by using the KMCMD algorithm  with the initial partition created using a numeric attribute.}
\label{Algorithmnumeric}
\begin{algorithmic}
\STATE {\bf Input-} Mixed dataset {\it T}, the
KMCMD algorithm, $k$ is the number of desired clusters.
\STATE Begin
\STATE {For $i^{th}$ numeric attribute}
\\
/*Assume that the $i^{th}$ attribute is normally distributed*/
\STATE 1-  Convert it to the standard  normal distribution using the following steps 
\STATE (i)- Calculate the mean $\mu$ for the $i^{th}$ numeric attribute
\STATE (ii)- Calculate the standard variation $\sigma$ for the $i^{th}$ numeric attribute.
\STATE (iii)- Calculate the z-score of each value ($x$) of the $i^{th}$ attribute using the formula $\frac{x - \mu}{\sigma}$.
\STATE 2- Find out $k$-1 values of z for which the area between each range is equal to $\frac{1}{k}$. We use -$\infty$ and +$\infty$ as the boundaries of extreme ranges. 
\STATE 3-  Divide the  dataset $T$ into $k$ clusters, depending on the range in which the attribute value falls.
\STATE 4- Use these clusters as the initial partition for the KMCMD algorithm and cluster the mixed data $T$ with the KMCMD algorithm.  
\STATE End     
\end{algorithmic}
\end{algorithm}

\begin{algorithm}[!ht]
\caption{$KMCMD\_Initial\_Categorical$ algorithm for creating clusters by using the KMCMD algorithm  with the initial partition created by using a categorical attribute.}
\label{Algorithmcategorical}
\begin{algorithmic}
\STATE {\bf Input-} Mixed dataset {\it T}, the KMCMD algorithm.
\STATE Begin
\STATE {For $i^{th}$ categorical attribute}
\STATE 1- Find unique attribute values,
\STATE 2- Create clusters based on attribute values such that data points having the same attribute value for the $i^{th}$ categorical attribute will be in the same cluster.  
\STATE 3-  Use the output of the last step as the initial partition for the KMCMD algorithm and cluster the mixed dataset {\it T}.
\\
\textcolor{black}{/*As the number of categorical values may be different from the value of $k$. The final number of clusters may be different than the value of $k$*/}
\STATE End      
\end{algorithmic}
\end{algorithm} 

\subsection{Initial partition using numeric attributes} 
Each numeric attribute is used to create initial clusters for one of the runs of the KMCMD algorithm. A numeric attribute is assumed to have a normal distribution \citep{kmeanclusterinitialization1}. Initial clusters are created such that the probability distributions of the attribute values are equal in each cluster. For \textit{k} clusters, \textit{k}-1 boundaries in the normal distribution graph are created so that the area between two adjacent boundaries is $\frac{1}{k}$. The extreme points -$\infty$ and +$\infty$  will also be used as boundaries along with the $k-1$ boundaries. For example, if we want to create three clusters from a numeric attribute, two points $z_1$ and $z_2$ are selected such that the area between -$\infty$ and $z_1$, $z_1$ and $z_2$, and $z_2$ and +$\infty$ is $\frac{1}{3}$. The data points are distributed in three clusters depending on the range in which the attribute value of a data point lies \textcolor{black}{to create initial partition.} Our proposed method is different from the method proposed by Khan and Ahmad (\citeyear{kmeanclusterinitialization1}) to compute the initial partition for k-means clustering algorithm as the latter computes the initial cluster centres by selecting a point in a given range such that the area under the curve in that range is divided equally. These centres are used to create the initial clusters for a run of the k-means clustering algorithm. However, as the normal distribution curve is not a straight line parallel to the horizontal axis, the probability distributions of the attribute values in the clusters are not equal. However, in the proposed method, the boundaries are computed in a way that the probability distributions of attribute values in the clusters are equal. Here, we would like to point out that we assume that  all the numeric attributes are normally distributed. Previous results suggest that the assumption of numeric attributes to follow normal distribution works well, in practice,  in finding initial partitions by k-means clustering \citep{kmeanclusterinitialization1,khan2003computing}. 

The KMCMD algorithm is run on the complete mixed dataset with the initial partition created by the numeric attribute resulting in a clustering result. 
The algorithm for creating clusters by using the KMCMD algorithm with the initial partition yielded using a
numeric attribute is presented in Algorithm \ref{Algorithmnumeric}.  

\subsection{Initial partition using categorical attributes} 
A categorical attribute consists of two or more categorical values. It has been shown that these attribute values can be used to create clusters \citep{Iam,Clusterensemble2005cs........9011H,kModeclusterinitilization1}. Following a similar methodology, we use the values of a categorical attribute to create an initial partition. For example, for an attribute with three attribute values $\alpha$, $\beta$ and $\gamma$, the points can be clustered in the three clusters based on these attribute values; \textcolor{black}{these clusters are then used as initial clusters.} Each categorical attribute is used to create the initial clusters for one of the runs. Khan and Ahmad (\citeyear{kModeclusterinitilization1}) use a similar approach to identify the initial partition for pure categorical datasets. They generally only use attributes with values that are equal to or less than the number of desired clusters to prevent a large number of distinct clustering strings from being created (a clustering string is a combination of all cluster labels for a data point, see Table \ref{tabensemb}). In contrast, the proposed approach has no such constraint; all the categorical attributes all used to create the initial partition. The \textit{initKmix} algorithm is for mixed datasets with two kinds of attributes, numeric and categorical. It is important that both types of attributes are treated equally. As we use each numeric attribute to create the initial partition in one of the runs of the KMCMD algorithm for mixed datasets, each categorical attribute should also be used in one of the runs of the KMCMD algorithm, \textcolor{black}{which is not the case in the approach proposed by Khan and Ahmad (\citeyear{kModeclusterinitilization1}).}

The KMCMD algorithm is run on the complete mixed dataset with the initial partition created by the categorical attribute to yield a clustering result. The method for creating clusters using the KMCMD algorithm for mixed datasets with initial partition created using a categorical attribute is presented in Algorithm
 \ref{Algorithmcategorical}.

\begin{table}[!ht]
	\centering
	\caption{An example of clustering results after step 1 of Algorithm 2. For a dataset with four attributes and five data points. A column represents a clustering result in a run. Four cluster labels will be generated for each data point. For example, in the first, second, third and fourth run, the first data point is given a cluster label a, b, b, and a respectively. } 
	\label{tabensemb}
	\begin{tabular}{|l|l|l|l|l|}
		\hline
		Data & First &Second & Third  &Fourth
		\\ point &run    & run  & run  &run
		\\
		\hline
		1&a & b & b & a
		\\2&b&a &b & a
		\\3&b& a& b & a
		\\4&b& b& a& b
		\\5&a& b &a& b
		\\
		\hline
	\end{tabular}
\end{table}

\subsection{Combining multiple clustering results} 
\label{sec:combining}
In the \textit{initKmix} algorithm, the KMCMD algorithm is run $m$ times to produce $m$ clustering results. An example of clustering results for different runs is presented in Table \ref{tabensemb}. These $m$ clustering results are combined to yield the initial partition.  A similar approach has been used by Khan and Ahmad (\citeyear{kmeanclusterinitialization1,kModeclusterinitilization1}), however, their method to combine $m$ clustering results can be quadratic with respect to the number of data points in the worst case. 

Several cluster ensemble algorithms have been developed to combine the results of multiple clustering results of a given dataset, resulting in superior clustering results \citep{DBLPghosh,Strehl,Topchy}. \textcolor{black}{Strehl and Ghosh \citep{Strehl} propose cluster ensemble algorithms that have a linear time complexity to the number of data points.} We use the following cluster ensemble algorithms to combine the $m$ clustering results to obtain the initial clusters. 

\begin{enumerate}[(i)]
\item HyperGraph partitioning algorithm - 
In this algorithm, the clustering ensemble problem is defined as the partitioning problem of a hypergraph, where hyperedges (a hyperedge is a generalization of an edge that can connect any set of vertices) represents clusters. The complexity of this method is \textit{$O(nkr)$} where $r$ is the number of runs of the clustering algorithm.
\item Meta-CLustering algorithm - 
In this algorithm, the cluster ensemble problem is considered to be the cluster correspondence problem. Groups of similar clusters are  identified and combined. The complexity of this method is \textit{$O(nk^2r^2)$}. 
\end{enumerate}

\subsection{The KMCMD algorithm}
As mentioned in Section \ref{sec:intro}, the k-means algorithm cannot directly be used to cluster mixed data because of the distance function in the objective function it optimizes. Ahmad and Dey (\citeyear{AmirLipika2007}) propose KMCMD that modifies the distance function of the standard k-means algorithm. In this paper, we use this algorithm in the \textit{initKmix} algorithms as a baseline to determine the initial partition. This initial partition will then be used with the KMCMD algorithm to produce the final clustering. 

In this algorithm, Ahmad and Dey (\citeyear{AmirLipika2007}) propose a distance measure for categorical attribute values. The weight of a numeric attribute, which represents the significance of the attribute, is also incorporated in the distance function to highlights its significance. A novel frequency-based definition of the cluster centre for categorical attributes is also proposed for a better representation of clusters. The modified distance function \citep{AmirLipika2007} computing the distance ($\psi(d_i,C_j)$) between the $i^{th}$ data point ($d_i$) and the $j^{th}$ cluster center ($C_j$) is given as 
 \begin{eqnarray}
 \label{eq:distfn}
\psi(d_i,C_j)= \underbrace{ \sum\limits_{t=1}^{m_r}(w^{r}_t(d^{r}_{it} - C^{r}_{jt}))^2}_{numeric} + \underbrace{ \sum\limits_{t=1}^{m_c}(\Omega(d^{c}_{it},C^{c}_{jt}))^2}_{categorical}
\end{eqnarray}  
where $w_t$ is the weight of the $t^{th}$ numeric attribute and 
$d^{r}_{it}$ is the value of the $t^{th}$ numeric attribute of the $i^{th}$ data point. $C^{r}_{jt}$ is the value of the $t^{th}$ numeric attribute of the $j^{th}$ cluster centre. $d^{c}_{it}$ is the value of the $t^{th}$ categorical attribute of the $i^{th}$ data point. $C^{c}_{jt}$ is the centre representation of the $j^{th}$ centre for the $t^{th}$ categorical attribute. $\Omega$ is the distance between a cluster centre and a data point for a categorical attribute.
As shown in Eq. (\ref{eq:distfn}), there are two terms, one each for computing the distance for the numeric and categorical attributes. For numeric attributes, the Euclidean distance with the weight of each numeric attribute is used. For categorical attributes, the frequency based definition for centre and co-occurrence based method to compute the distance between two attribute values is used.

This distance measure does not take the distance between two attribute values of a categorical attribute as 0 or 1 (Hamming distance), rather it computes the distance between two values of an attribute from the dataset. The distance between two attribute values $\alpha$ and $\beta$ with respect to the other attribute is computed by using the following formula 
\begin{equation}
\label{eq:catdist}
max\mid p(f{\mid}{\alpha})+p({\sim}f{\mid}\beta)\mid - 1
\end{equation} 

where $p(f{\mid}{\alpha})$ represents the probability for data point with attribute value $\alpha$ having other attribute values belonging to a subset of classes $f$, whereas $p({\sim}f{\mid}{\beta})$ represents the probability for data points with attribute value $\beta$ having other attribute values not belonging to $f$. Out of many subsets of classes, a subset maximizing the value in Eq. (\ref{eq:catdist}) is selected. The distance between two values of an attribute is computed with respect to all the other attributes and the average is taken as the distance between these two attribute values.

The distance algorithm  does not take the significance of the numeric attributes as equal but computes the significance of a numeric attribute from the dataset. A numeric attribute is discretized; the new attribute is treated as a categorical attribute.  The average of the distances of all the pairs of attribute values is taken as the weight of the numeric attribute. The discretization of numeric attributes is undertaken only to compute the weight of the numeric attributes. The clustering is performed with numeric attributes. The complete clustering algorithm is presented in Algorithm \ref{Algorithmmixed}.

\begin{algorithm}[!ht]
\caption{The KMCMD algorithm \citep{AmirLipika2007}.}
\label{Algorithmmixed}
\begin{algorithmic}
\STATE {\bf Input-} Mixed dataset {\it T}, the number of attributes $m$, the number of numeric attributes $m_r$ and the number of categorical attributes $m_c$. $k$ the number of desired clusters.
\STATE Begin
\STATE For all numeric attributes
\FOR{i=1...$m_r$}
\STATE 1- {Normalize all numeric attributes.}
\STATE 2- {Discretize all numeric attributes.} 
\\
/*It will be considered as categorical attribute.*/
\ENDFOR 
\STATE \textcolor{black}{ For all categorical attributes (categorical attributes in the original dataset and discretized numeric attributes)}
\FOR{i=1...$m$}
\STATE     1- Compute the distance between every pair of attribute values by using  the co-occurrence of the pair with respect to other attributes  (using Eq. (\ref{eq:catdist})).
\ENDFOR 
\STATE For all numeric attributes
\FOR{i=1...$m_r$}
\STATE  1- Compute the distance between every pair of discretized attribute values.
\STATE  2- The average of all the distances (between each pair of attribute values) is taken as the weight of the numeric attribute.
\ENDFOR
\\
/*Following steps are for clustering*/
\STATE (1) Take the original dataset that has normalized numeric attributes and categorical attributes, assign the data points to $k$ clusters randomly. 
\STATE (2) Repeat steps $A$ and $B$ 
\STATE (A) Calculate the centres of the clusters. 
\\
/*For a numeric attribute, the mean is used to define a cluster centre whereas for a categorical attribute. A frequency-based measure is used to define a cluster centre.*/
\STATE (B) Each data point is assigned to its nearest cluster using the distance measure defined in Eq. (\ref{eq:distfn}).
\STATE Until no data point changes cluster membership or the predefined number of iterations is reached.
\STATE End 
\end{algorithmic}
\end{algorithm}

\subsection{Computational complexity} The \textit{initKmeans} algorithm  run the KMCMD algorithm \citep{AmirLipika2007} $m$ times for a dataset (with $n$ data points) to create $m$ clustering results corresponding to each $m$ attributes. The complexity of the KMCMD algorithm is $O(m^2n + m^2S^3 + pn(km_r + km_cS))$ where $p$ is the number of iterations and $S$ is the average number of distinct categorical values. Hence, for $m$ number of runs, the complexity is $O(m(m^2n + m^2S^3 + pn(km_r + km_cS)))$. Then these $m$ clustering results are combined to obtain a clustering result. These results are combined using  HyperGraph partitioning algorithm  (complexity \textit{$O(nkm)$} ) or Meta-CLustering algorithm (complexity \textit{$O(nk^2m^2)$}) \citep{Strehl}. This clustering result is used as an initial partition to run the KMCMD algorithm. Hence, the total complexity when clustering results are combined with: 

HyperGraph partitioning algorithm is

\begin{equation*}
\begin{split}
 O(((m+1)(m^2n + m^2S^3 + pn(km_r + km_cS))) + O(nkm))\\
\implies  O((m(m^2n + m^2S^3 + pn(km_r + km_cS))) + (nkm))
\end{split}
\end{equation*} 

Meta-CLustering algorithm is
\[O(m(m^2n + m^2S^3 + pn(km_r + km_cS)) + O(nk^2m^2))\]

In both the cases, the time complexity is linear to the number of data points. In the experiment, both ensemble algorithms are run and the clustering results with better normalized mutual information is selected. As both the ensemble algorithms are linear to the number of data points, the total complexity of the clustering algorithm will remain linear to the number of data points. 

\section{Experiments and Results} 
\label{sec:experiments}
We implemented the \textit{initKmix} and the KMCMD algorithm \citep{AmirLipika2007} using Java JDK 1.8.
To perform cluster ensemble step, the Octave implementations of the cluster ensemble  algorithms were used \citep{clusterensemblelink}. A minor modification was made to the clustering ensemble implementation such that the method only considered HyperGraph Partitioning Algorithm and Meta-CLustering Algorithm based on the maximum average normalized mutual information \citep{Strehl}.  
 The \textit{initKmix} algorithm was first tested on a simulated mixed dataset, then on four pure categorical datasets and five mixed datasets downloaded from UCI repository \citep{Dua:2019}. All these datasets have predefined class and class labels, which were taken as ground truth. The number of the desired clusters was set to the number of the classes.  The clustering accuracy was computed against the ground truth. Each cluster was mapped to a distinct class so that the following measure \citep{AmirLipika2007} had the maximum value;
\begin{eqnarray}
AC = \frac{\sum\limits_{i=1}^KG_i}{n} 
\end{eqnarray} 
Where $G_i$ is the number of data points correctly assigned to a class. 

This measure is called clustering accuracy ($AC$) and has been used to compare the clustering results \citep{AmirLipika2007}. Two measures derived from $AC$, $\overline{AC}$ (for the average clustering accuracy) and $SD$ (for the standard deviation of clustering accuracy), were also used to present the results of clustering methods with random initial clusters;  The average clustering accuracy for $T$  runs is defined in the following way;

\begin{eqnarray}
\overline{AC} = \frac{\sum\limits_{i=1}^TAC_i}{T} 
\end{eqnarray} 
where $AC_i$ is the clustering accuracy in the $i^{th}$ run.

 The standard deviation of the clustering accuracy for $T$  runs is computed in the following way;
 \begin{equation}\label{key}
SD_{AC} =\sqrt{\frac{\sum\limits_{i=1}^T(AC_i - \overline{AC})^2}{T})} 
 \end{equation}
Where $AC_i$ is the clustering accuracy in the $i^{th}$ run.

\textcolor{black}{Two other clustering  performance measures, Rand index ($RI$) and Adjusted Rand index ($ARI$) \citep{Rand} were also employed to compute the clustering performance. $RI$ represents the frequency of occurrence of agreements between two clustering over the total pairs of data points. $RI$ is defined by following expression;}
\begin{eqnarray}
RI = \frac{a+b}{\binom{n}{2}}
\end{eqnarray} 
\textcolor{black}{where $a$  in the the number of  pairs of data points belong to the same cluster across two different clustering results and the $b$  is the number of  pairs of data points are in different clusters across two different clustering results. Classes were taken as one clustering and the clustering is compared with it.}

\textcolor{black}{$ARI$ is the corrected-for-chance version of the Rand index.
$ARI$ is defined by following expression;}

\begin{eqnarray}
ARI = \frac{ \sum_{ij} \binom{n_{ij}}{2} - [\sum_i \binom{a_i}{2} \sum_j \binom{b_j}{2}] / \binom{n}{2} }{ \frac{1}{2} [\sum_i \binom{a_i}{2} + \sum_j \binom{b_j}{2}] - [\sum_i \binom{a_i}{2} \sum_j \binom{b_j}{2}] / \binom{n}{2} }
\end{eqnarray} 

 \textcolor{black}{$n_{ij}$ denotes the number of data points common between  $i^{th}$ cluster of clustering and $j^{th}$ class, $a_{i}$ refers the number of data points in $i^{th}$ cluster and $b_{i}$ denotes the number of data points in $j^{th}$ class.}
 
 \textcolor{black}{Similar to $\overline{AC}$, for many runs of the KMCMD algorithm,  the average values of $RI$ ($\overline{RI}$) and $ARI$ ($\overline{ARI}$) were computed. Standard deviation of $RI$ ($SD_{RI}$) and $ARI$ ($SD_{ARI}$) were also used to compute the performance of clustering algorithms.}
 
The higher values of performance measures ($AC$, $RI$ and $ARI$) suggest better clustering results. The maximum possible value for these performance measures is $1$ that represents  that the data clustering and data classes are exactly the same. Similarly, for the KMCMD algorithm with random initial partitions, high values of ($\overline{AC}$, $\overline{RI}$ and $\overline{ARI}$) are desired. $SD$ represents the inconsistency of clustering results with a different initialization in each run of the KMCMD algorithm. The low values of $SD$ ($SD_{AC}$, $SD_{RI}$ and $SD_{ARI}$) point to the robustness of the algorithm for different initial partitions. 

We carried out the two types of experiments. First, we compared the \textit{initKmix} algorithm against the random initial partition method. The KMCMD algorithm \citep{AmirLipika2007} was run 50 times with a random initial partition method and average clustering performances are presented. We ran the \textit{initKmix} algorithm just once for a dataset to obtain the initial partition, and then the KMCMD algorithm \citep{AmirLipika2007} was applied with this initial partition to yield the clustering result.  As this approach yields identical results every time, the $SD$ of clustering results for the \textit{initKmix} algorithm is $0$ for any dataset. 

 Second, the KMCMD algorithm \citep{AmirLipika2007} with \textit{initKmix} algorithm was also compared with other clustering algorithms using $AC$ performance measure. Results for these clustering algorithms were taken from the published papers. Majority of papers on mixed data clustering methods use similar datasets and performance measure ($AC$). Therefore, we also use those frequently used datasets and the performance measure to facilitate comparison of various clustering algorithms with the KMCMD algorithm with the \textit{initKmix} algorithm. 
 
 Now, we show the results on a mixed simulated dataset, followed by results on pure categorical and mixed datasets.

 \subsection{Simulated mixed data}\textcolor{black}{We compared the performance of \textit{initKmix} algorithm against random initial partition on a simulated mixed dataset. Following \textit{clustMixType} package \citep{Rclusmix} of R \citep{Rsoft}, a mixed dataset was generated. The dataset had four attributes, two of them were categorical, whereas the other two were numeric. Each categorical attribute had two categories, $A$ and $B$ for the first categorical attribute and $X$ and $Y$ for the second attribute. The numeric attributes were created using normal distribution. There were 400 data points divided equally into four clusters.  First attribute values of cluster, $C_1$, were created by selecting attribute categories $A$ and $B$ randomly with 0.9 and 0.1 probability respectively. Similarly the second attribute values of $C_1$, were created by selecting attribute categories $X$ and $Y$. Third attribute values were created using normal distribution with $\mu$ = -5 and $\sigma$ = 1. The fourth attribute values generated using the same method as of attribute three. All the remaining three clusters ($C_2$, $C_3$, $C_4$) were created with similar procedures but with different parameters. The properties of each cluster are presented in Table \ref{simulateddata}. We ran KMCMD algorithm with random initial partition 50 times and studied the final clustering results. Final clustering results were inconsistent in different runs. Mostly, data points were clustered in two, three or four clusters. Examples of these clustering results are given in Tables \ref{simulateddatarand1} - \ref{simulateddatarand3}. } \textcolor{black}{It is to be noted that in two different clustering runs there is no direct correspondence between final cluster labels. For example, $C_{f1}$ (cluster label) in one instance of clustering may not be $C_{f1}$ (cluster label) in the other instance of clustering. }
 
 \textcolor{black}{Clustering results (Table \ref{simulateddatainitial}) demonstrate that in contrast to KMCMD with the random partition,  KMCMD with the \textit{initKmix} algorithm was able to identify four clusters for the dataset accurately ($AC$ = 0.908, $RI$ = 0.916 and $ARI$ = 0.775). We would like to emphasise that \textit{initKmix} algorithm generates only one clustering result. This further highlights that that application of \textit{initKmix} algorithm produces accurate and consistent clustering results.}
 \begin{table}[!ht]
\centering
\caption{\textcolor{black}{Description of the clusters of the simulated dataset.}}
  \begin{tabular}{|l|l|l|l|l|}
  \hline
  \multirow{2}{*}{Cluster}   &
      \multicolumn{4}{c|}{Attributes } \\
      \cline{2-5}
     &Categorical-1 & Categorical-2& Numeric-1&Numeric-2 \\
     \hline      
    $C_1$ & A (90\%), B (10\%) & X (90\%), Y (10\%) &$\mathcal{N}(-5,1)$ & $\mathcal{N}(-5, 1)$ \\
    \hline
    $C_2$ & A (90\%), B (10\%) & X (90\%), Y (10\%) &$\mathcal{N}(5, 1)$ & $\mathcal{N}(5, 1)$  \\
    \hline
    $C_3$ & A (10\%), B (90\%) & X (10\%), Y (90\%) &$\mathcal{N}(-5, 1)$ & $\mathcal{N}(-5, 1)$ \\
    \hline
    $C_4$ & A (10\%), B (90\%) & X (90\%), Y (10\%) &$\mathcal{N}(5, 1)$ & $\mathcal{N}(5, 1)$  \\
    \hline
  \end{tabular}
  \label{simulateddata}
\end{table}
 
 \begin{table}[!ht]
\centering
\caption{\textcolor{black}{One of the clustering results by the KMCMD algorithm with random initial partition. Most of the data points are in two clusters.} }
  \begin{tabular}{|c|c|c|c|c|}
   \hline
      \multirow{2}{*}{Original cluster}   &
      \multicolumn{4}{c|}{Final clusters } \\
      \cline{2-5}
    &$C_{f1}$ & $C_{f2}$ & $C_{f3}$ &$C_{f4}$  \\
     \hline      
    $C_1$ & 10 & 0 &		90&0 \\
    \hline
    $C_2$ & 0&90&0&	10 \\
    \hline
     $C_3$ &7&0&		93&0 \\
    \hline
    $C_4$& 0&90&0&	10 \\
    \hline
  \end{tabular}
  \label{simulateddatarand1}
\end{table}
 
 \begin{table}[!ht]
\centering
\caption{\textcolor{black}{One of the clustering results by the KMCMD algorithm with random initial partition. All the data points are in three clusters.}}
  \begin{tabular}{|c|c|c|c|}
   \hline
      \multirow{2}{*}{Original cluster}   &
      \multicolumn{3}{c|}{Final clusters } \\
      \cline{2-4}
    &$C_{f1}$ & $C_{f2}$ & $C_{f3}$  \\
     \hline      
    $C_1$ &0& 82 &	18 \\
    \hline
    $C_2$ & 18 &82 & 0 \\
    \hline
     $C_3$ & 0 & 0 & 100\\
    \hline
    $C_4$& 100&0&0\\
    \hline
  \end{tabular}
  \label{simulateddatarand2}
\end{table}

\begin{table}[!ht]
\centering
\caption{\textcolor{black}{One of the clustering results by the KMCMD algorithm with random initial partition. The clustering correctly produced four almost equal sized clusters.}}
  \begin{tabular}{|c|c|c|c|c|}
   \hline
      \multirow{2}{*}{Original cluster}   &
      \multicolumn{4}{c|}{Final clusters } \\
      \cline{2-5}
    &$C_{f1}$ & $C_{f2}$ & $C_{f3}$ &$C_{f4}$  \\
     \hline      
    $C_1$ & 0&	82	& 0&	18 \\
    \hline
    $C_2$ & 11 & 0 & 89& 0 \\
    \hline
     $C_3$ &0&0&0&100\\
    \hline
    $C_4$&91&0&		9&0 \\
    \hline
  \end{tabular}
  \label{simulateddatarand3}
\end{table}

\begin{table}[!ht]
\centering
\caption{\textcolor{black}{Clustering results by the KMCMD algorithm with the \textit{initKmix} algorithm. The final clustering has four quite accurate clusters.}}
  \begin{tabular}{|c|c|c|c|c|}
   \hline
      \multirow{2}{*}{Original cluster}   &
      \multicolumn{4}{c|}{Final clusters } \\
      \cline{2-5}
    &$C_{f1}$ & $C_{f2}$ & $C_{f3}$ &$C_{f4}$  \\
     \hline      
    $C_1$ & 90&0&0&			10 \\
    \hline
    $C_2$ &0&89&	11&0 \\
    \hline
     $C_3$ &7&0&0&	93 \\
    \hline
    $C_4$& 0& 9	&91 &0 \\
    \hline
  \end{tabular}
  \label{simulateddatainitial}
\end{table}
 
\subsection{Categorical datasets}We carried out an experiment with four categorical datasets; Soybean-small, Vote, Breast cancer and Mushroom \citep{Dua:2019}. Information on these datasets is provided in Table \ref{Categoricaldata}.  We apply the KMCMD algorithm on those datasets. The numeric part of the distance measure presented in Eq. (2) will be zero in this case. The  clustering results of the KMCMD algorithm with \textit{initKmix} and the KMCMD algorithm with the random partition method using different measures ($AC$, $RI$ and $ARI$) using are presented in Tables \ref{CategoricalIntial} -  \ref{CategoricalIntialadjrand}. \textcolor{black}{Unpaired t-test with 95\% confidence interval \citep{statsbook} was carried out to compare the performance of the KMCMD algorithm with the \textit{initKmix} algorithm and the KMCMD algorithm with the random partition method. The calculations indicated that for all categorical datasets the \textit{initKmix} algorithm performed statistically better than the random partition method. } 

The performance ($AC$) of the KMCMD algorithm with  \textit{initKmix} algorithm was also compared with k-modes algorithm with different state-of-the-art initialization methods; including Wu's initialization \citep{InitializationWu},  Cao's initialization \citep{InitializationCao}, Khan and Ahmad's initialization\citep{kModeclusterinitilization1} and Ini\_Entropy \citep{iniJiang2016}. The  KMCMD algorithm with  the \textit{initKmix} algorithm was also compared with the Fuzzy k-modes clustering algorithm \citep{Fuzzykmode} and CRAFTER \citep{RFclusters} algorithm.
The results of various clustering methods are presented in Table \ref{Categoricalallcllus}. The results for the other clustering algorithms were taken from existing papers \citep{kModeclusterinitilization1,iniJiang2016,RFclusters,ZHU2018230}. Except for the Soybean-small dataset, the combination of the KMCMD algorithm and \textit{initKmix} outperformed the other clustering methods. For the Soybean-small dataset, the KMCMD algorithm with the \textit{initKmix} algorithm performed similarly or better than the other clustering methods. 
\begin{table}[!ht]
\centering
\caption{Description of the categorical datasets used in the experiments}
  \begin{tabular}{|l|l|l|l|}
    \hline
    Dataset&Number of & Number of & Number of \\
           & data points&attributes&classes \\
     \hline      
    Soybean-small & 47 & 35 & 4\\
    \hline
    Vote & 435 & 16 & 2 \\
    \hline
    Breast cancer& 699 & 9 & 2 \\
    \hline
    Mushroom & 8124& 22& 2 \\
    \hline
  \end{tabular}
  \label{Categoricaldata}
\end{table}

\begin{table}[!ht]
\centering
\caption{The clustering results ($AC$) of the KMCMD algorithm with \textit{initKmix} and the KMCMD algorithm with the random initial partition method for categorical datasets. The better clustering result is shown in bold.}
 % \begin{tabular}{|p{20mm}|p{20mm}|p{20mm}|p{20mm}|p{20mm}|}
 \begin{tabularx}{\linewidth}{|X|X|X|X|X|}
    \hline
    \multirow{2}{*}{Dataset} & \multicolumn{2}{>{\centering\setlength\hsize{2\hsize} }X|}{The KMCMD algorithm  with \textit{initKmix}} & \multicolumn{2}{>{\centering\setlength\hsize{2\hsize} }X|}{The KMCMD algorithm with the random initial partition} \\ 
    %      \multicolumn{2}{c|}{The KMCMD algorithm  with \textit{initKmix}} & 
%      \multicolumn{2}{c|}{The KMCMD algorithm with the random initial partition} \\
   & $\overline{AC}$ & $SD_{AC}$ & $\overline{AC}$ & $SD_{AC}$  \\
    \cline{2-5}%\hline
    Soybean-small & \bf{1} & 0 & 0.967 & 0.079 \\
    \hline
    Vote & \bf{0.873} & 0& 0.871 & 0.002 \\
    \hline
    Breast cancer& \bf{0.974} & 0 & 0.965 & 0.014 \\
    \hline
    Mushroom & \bf{0.894} & 0 & 0.822 & 0.124 \\
    \hline
  \end{tabularx}
  \label{CategoricalIntial}
\end{table}

\begin{table}[!ht]
\centering
\caption{\textcolor{black}{The clustering results ($RI$) of the KMCMD algorithm with \textit{initKmix} and the KMCMD algorithm with the random initial partition method for categorical datasets. The better clustering result is shown in bold.}}
 % \begin{tabular}{|p{20mm}|p{20mm}|p{20mm}|p{20mm}|p{20mm}|}
 \begin{tabularx}{\linewidth}{|X|X|X|X|X|}
    \hline
    \multirow{2}{*}{Dataset} & \multicolumn{2}{>{\centering\setlength\hsize{2\hsize} }X|}{The KMCMD algorithm  with \textit{initKmix}} & \multicolumn{2}{>{\centering\setlength\hsize{2\hsize} }X|}{The KMCMD algorithm with the random initial partition} \\ 
    %      \multicolumn{2}{c|}{The KMCMD algorithm  with \textit{initKmix}} & 
%      \multicolumn{2}{c|}{The KMCMD algorithm with the random initial partition} \\
   & $\overline{RI}$ & $SD_{RI}$ & $\overline{RI}$ & $SD_{RI}$  \\
    \cline{2-5}%\hline
    Soybean-small & \bf{1} & 0 & 0.953 &0.083\\
    \hline
    Vote & \bf{0.779} & 0 & 0.775 & 0.003  \\
    \hline
    Breast cancer& \bf{0.950} & 0 &0.931 & 0.016 \\
    \hline
    Mushroom & \bf{0.811} & 0 & 0.708 & 0.147 \\
    \hline
  \end{tabularx}
  \label{CategoricalIntialrand}
\end{table}

\begin{table}[!ht]
\centering
\caption{\textcolor{black}{The clustering results ($ARI$) of the KMCMD algorithm with \textit{initKmix} and the KMCMD algorithm with the random initial partition method for categorical datasets. The better clustering result is shown in bold.}}
 % \begin{tabular}{|p{20mm}|p{20mm}|p{20mm}|p{20mm}|p{20mm}|}
 \begin{tabularx}{\linewidth}{|X|X|X|X|X|}
    \hline
    \multirow{2}{*}{Dataset} & \multicolumn{2}{>{\centering\setlength\hsize{2\hsize} }X|}{The KMCMD algorithm  with \textit{initKmix}} & \multicolumn{2}{>{\centering\setlength\hsize{2\hsize} }X|}{The KMCMD algorithm with the random initial partition} \\ 
    %      \multicolumn{2}{c|}{The KMCMD algorithm  with \textit{initKmix}} & 
%      \multicolumn{2}{c|}{The KMCMD algorithm with the random initial partition} \\
  & $\overline{ARI}$ & $SD_{ARI}$ & $\overline{ARI}$ &  $SD_{ARI}$ \\
    \cline{2-5}%\hline
    Soybean-small & \bf{1} & 0 & 0.873 & 0.091 \\
    \hline
    Vote & \bf{0.557} & 0&0.550  &0.004  \\
    \hline
    Breast cancer& \bf{0.899} & 0 &0.861  & 0.017  \\
    \hline
    Mushroom & \bf{0.622}& 0 & 0.415 & 0.163 \\
    \hline
  \end{tabularx}
  \label{CategoricalIntialadjrand}
\end{table}

\begin{table}[!ht]
\scriptsize
\centering
\caption{Results ($AC$) for various clustering algorithms for categorical datasets.``-'' denotes results for a given algorithm that could not be obtained from the literature. The best clustering results are shown in bold.}
  \begin{tabular}{|p{12mm}|p{12mm}|p{12mm}|p{12mm}|p{12mm}|p{12mm}|p{12mm}|p{12mm}|p{12mm}|}
    \hline
    Dataset&The KMCMD algorithm with \textit{initKmix}&k-modes \citep{Huang97afast} with random initialization&k-modes with Wu's initialization \citep{InitializationWu}&k-modes  with Cao's initialization \citep{InitializationCao}&k-modes  with Khan and Ahmad's initialization \citep{kModeclusterinitilization1}&k-modes  with Ini\_Entropy initialization \citep{iniJiang2016}&CRAFTER \citep{RFclusters}& \textcolor{black}{Fuzzy k-modes clustering \citep{Fuzzykmode}}\\
    \hline
    Soybean- small & \bf{1} & 0.864 & \bf{1.000} & \bf{1.000} &0.957&\bf{1.000}&\bf{1.000} &\textcolor{black}{0.824}\\
    \hline
    Vote & \bf{0.873} & 0.842& - & -& 0.850&0.869&0.856&\textcolor{black}{0.862}\\
    \hline
    Breast cancer& \bf{0.974} & 0.836 & 0.911 & 0.911&0.913&0.933&-&- \\
    \hline
    Mushroom & \bf{0.894} & 0.875 & 0.875 & 0.875&0.882&0.888&0.774&\textcolor{black}{0.723}\\
    \hline
  \end{tabular}
  \label{Categoricalallcllus}
\end{table}

\subsection{Mixed datasets} The following five mixed datasets \citep{Dua:2019} were used in the experiments: Acute Inflammations, Heart (Statlog), Heart (Cleveland),  Australian credit and German credit. Table \ref{mixeddatainfo} displays information on these datasets.  The  clustering results by the KMCMD algorithm with the \textit{initKmix} algorithm and the KMCMD algorithm with the random partition method using different measures ($AC$, $RI$ and $ARI$) using are presented in Tables \ref{mixeddatarandom} - \ref{mixeddatarandomadrand}. \textcolor{black}{Unpaired t-test with 95\% confidence interval (sample sizes were 50) was carried out to compare the performance of KMCMD algorithm with \textit{initKmix} and the KMCMD algorithm with random inital partition. Results suggest that except Australian credit dataset for $AC$ performance measure (Table \ref{mixeddatarandom}),  the \textit{initKmix} algorithm performed statistically better than the random partition method for all other datasets on all the performance measures. For Australian credit dataset, there is no statistically significant difference \textit{initKmix} between the performances of two methods. However, \textit{initKmix} algorithm produced consistent clustering. }

The performance of the KMCMD algorithm with the  \textit{initKmix} algorithm was also compared using $AC$ performance measure with k-prototypes \citep{Huang1997} with random initialization, k-prototypes \citep{Huang1997} with the Ji et al.
		\citep{ji2015novel} initialization method, Similarity-based Agglomerative clustering (SBAC) \citep{LiBiswasclustering}, Object-cluster similarity metric (OCIL) algorithm \citep{CHEUNG20132228} and fuzzy k-prototypes clustering \citep{ji2012fuzzy}.
The results of various clustering methods are presented in Table \ref{mixeddataallcluster}. The results for the other clustering algorithms were taken from published papers \citep{ji2012fuzzy,CHEUNG20132228,DU201746}. Except for the German credit dataset, the KMCMD algorithm with \textit{initKmix} performed better than the other clustering algorithms. For the German credit dataset, the OCIL algorithm performed better than the KMCMD algorithm with \textit{initKmix}.

\begin{table}[!ht]
\centering
\caption{Description of the mixed datasets used in the experiments}
  \begin{tabular}{|l|l|l|l|l|}
    \hline
    Dataset&Number of & Number of & Number of &Number of\\
           &data points&categorical&numeric&classes \\
           & &attributes&attributes& \\
     \hline      
    Acute Inflammations & 120 & 5 & 1&2\\
    \hline
    Heart (Statlog) & 270&7 & 6 & 2 \\
    \hline
    Heart (Cleveland)& 303&6 &7& 2 \\
    \hline
    Australian credit & 690&8&6& 2 \\
    \hline
    German credit&1000&13&7&2\\
    \hline
  \end{tabular}
  \label{mixeddatainfo}
\end{table}

\begin{table}[!ht]
\centering
\caption{The clustering results ($AC$) of the KMCMD algorithm with \textit{initKmix} and the KMCMD algorithm with the random initial partition method for mixed datasets. The better clustering result is shown in bold.}
 \begin{tabularx}{\linewidth}{|X|X|X|X|X|}
    \hline
    \multirow{2}{*}{Dataset} & \multicolumn{2}{>{\centering\setlength\hsize{2\hsize} }X|}{The KMCMD algorithm  with \textit{initKmix}} & \multicolumn{2}{>{\centering\setlength\hsize{2\hsize} }X|}{The KMCMD algorithm with the random initial partition} \\ 
    
%   \begin{tabular}{|p{20mm}|p{20mm}|p{20mm}|p{20mm}|p{20mm}|}
%     \hline
%     \multirow{2}{*}{Dataset} &
%       \multicolumn{2}{c|}{The KMCMD algorithm with \textit{initKmix}} & 
%       \multicolumn{2}{c|}{The KMCMD algorithm with the random initial partition} \\
    & $\overline{AC}$ & $SD_{AC}$ & $\overline{AC}$ & $SD_{AC}$ \\
    \cline{2-5}%hline
    Acute infammation & \bf{0.823}& 0& 0.762& 0.125 \\
    \hline
    Heart (Statlog) & \bf{0.817} & 0& 0.802 &0.010 \\
    \hline
    Heart (Cleveland)& \bf{0.841} & 0 & 0.834 &0.005 \\
    \hline
    Australian credit & \bf{0.858} & 0 & 0.829& 0.118 \\
    \hline
    German credit & \bf{0.683} & 0 & 0.678& 0.004 \\
    \hline
  \end{tabularx}
  \label{mixeddatarandom}
\end{table}

\begin{table}[!ht]
\centering
\caption{\textcolor{black}{The clustering results ($RI$) of the KMCMD algorithm with \textit{initKmix} and the KMCMD algorithm with the random initial partition method for mixed datasets. The better clustering result is shown in bold.}}
 \begin{tabularx}{\linewidth}{|X|X|X|X|X|}
    \hline
    \multirow{2}{*}{Dataset} & \multicolumn{2}{>{\centering\setlength\hsize{2\hsize} }X|}{The KMCMD algorithm  with \textit{initKmix}} & \multicolumn{2}{>{\centering\setlength\hsize{2\hsize} }X|}{The KMCMD algorithm with the random initial partition} \\ 
    
%   \begin{tabular}{|p{20mm}|p{20mm}|p{20mm}|p{20mm}|p{20mm}|}
%     \hline
%     \multirow{2}{*}{Dataset} &
%       \multicolumn{2}{c|}{The KMCMD algorithm with \textit{initKmix}} & 
%       \multicolumn{2}{c|}{The KMCMD algorithm with the random initial partition} \\
    & $\overline{RI}$ & $SD_{RI}$ & $\overline{RI}$ &  $SD_{RI}$\\
    \cline{2-5}%hline
    Acute inflammation & \bf{0.709}& 0&0.639 & 0.118 \\
    \hline
    Heart (Statlog) & \bf{0.679} & 0& 0.665 &0.013 \\
    \hline
    Heart (Cleveland)& \bf{0.728} & 0 & 0.719&0.005 \\
    \hline
    Australian credit & \bf{0.756} & 0 &0.716 & 0.127 \\
    \hline
    German credit & \bf{0.567} & 0 &0.563& 0.003 \\
    \hline
  \end{tabularx}
  \label{mixeddatarandomrand}
\end{table}

\begin{table}[!ht]
\centering
\caption{\textcolor{black}{The clustering results ($ARI$) of the KMCMD algorithm with \textit{initKmix} and the KMCMD algorithm with the random initial partition method for mixed datasets. The better clustering result is shown in bold.}}
 \begin{tabularx}{\linewidth}{|X|X|X|X|X|}
    \hline
    \multirow{2}{*}{Dataset} & \multicolumn{2}{>{\centering\setlength\hsize{2\hsize} }X|}{The KMCMD algorithm  with \textit{initKmix}} & \multicolumn{2}{>{\centering\setlength\hsize{2\hsize} }X|}{The KMCMD algorithm with the random initial partition} \\ 
    
%   \begin{tabular}{|p{20mm}|p{20mm}|p{20mm}|p{20mm}|p{20mm}|}
%     \hline
%     \multirow{2}{*}{Dataset} &
%       \multicolumn{2}{c|}{The KMCMD algorithm with \textit{initKmix}} & 
%       \multicolumn{2}{c|}{The KMCMD algorithm with the random initial partition} \\
    & $\overline{ARI}$ & $SD_{ARI}$ & $\overline{ARI}$ &  $SD_{ARI}$\\
    \cline{2-5}%hline
    Acute inflammation & \bf{0.415}& 0&0.273 & 0.182 \\
    \hline
    Heart (Statlog) & \bf{0.358} & 0& 0.331 &0.021 \\
    \hline
    Heart (Cleveland)& \bf{0.456} & 0 &0.438  &0.009 \\
    \hline
    Australian credit & \bf{0.512} & 0 &0.432 & 0.217 \\
    \hline
    German credit & \bf{0.0519} & 0 &0.0415 & 0.006 \\
    \hline
  \end{tabularx}
  \label{mixeddatarandomadrand}
\end{table}

\begin{table}[!ht]
\centering
\caption{Results ($AC$) for various clustering algorithms for mixed datasets. ``-'' denotes results for a given algorithm that could not be obtained from the literature. The best clustering result is shown in bold.}
  \begin{tabular}{|p{15mm}|p{15mm}|p{15mm}|p{15mm}|p{15mm}|p{15mm}|p{15mm}|}
    \hline
    Dataset &The KMCMD algorithm with \textit{initKmix}&k-prototypes \citep{Huang1997} with random initialization&k-prototypes \citep{Huang1997} with Ji et al.
		\citep{ji2015novel} initialization method &Similarity-based Agglomerative clustering (SBAC) \citep{LiBiswasclustering}&Object-cluster similarity metric (OCIL) algorithm \citep{CHEUNG20132228}&\textcolor{black}{Fuzzy k-prototypes \cite{ji2012fuzzy}} \\
	\hline	
    Acute inflammation & \bf{0.823}& 0.610& -& 0.508&-&\textcolor{black}{0.710} \\
    \hline
    Heart (Statlog) & \bf{0.817} & 0.770& - &-&0.814&- \\
    \hline
    Heart (Cleveland)& \bf{0.841} & 0.772 & 0.808 &0.752&0.831&\textcolor{black}{0.835} \\
    \hline
    Australian credit & \bf{0.858} & 0.738& 0.800&0.600& 0.757&\textcolor{black}{ 0.838}\\
    \hline
    German credit & 0.683 & 0.671 & -& -&\bf{0.695}&- \\
    \hline
  \end{tabular}
  \label{mixeddataallcluster}
\end{table}

\subsection{Discussion}\textcolor{black}{ Wilcoxon Signed-Ranks test \citep{Wisconsintest} with 95\% confidence level  was carried out to compare the performance of the KMCMD algorithm with the \textit{initKmix} algorithm against KMCMD algorithm  with random initial partition over all the nine datasets. The test suggests that the  KMCMD algorithm with the \textit{initKmix} algorithm  significantly better than KMCMD algorithm  with the random initial partition method.}

 The KMCMD algorithm with \textit{initKmix} perform similar to or better than the other state-of-the-art clustering algorithms for categorical datasets. Some of these clustering methods use different initialization methods \citep{kModeclusterinitilization1,InitializationWu,InitializationCao}, and the better clustering results with the \textit{initKmix} algorithm suggests that the it produces good initial partition. The similar behaviour is observed for mixed datasets. The proposed approach has the best performance across clustering methods for mixed datasets for four out of five datasets. One of these clustering methods \citep{ji2015novel} uses an initialization method, but the better clustering results point to the superiority of the \textit{initKmix} algorithm in creating an initial partition.

In \textit{initKmix} algorithm, we get initial partition of a dataset for the KMCMD algorithm after combining many clustering results. It is possible that this initial partition can be used as the final clustering result. However, the goal of this paper is to study the performance of a KMCMD algorithm with initial partition created by \textit{initKmix} algorithm. 

\textcolor{black}{The results suggest that the KMCMD algorithm  with the \textit{initKmix} algorithm produces accurate clustering  for both categorical and mixed datasets. The \textit{initKmix} algorithm generates the accurate initial partition, which in turn improves the performance of KMCMD in comparison to random initial partition. The accurate and diverse clustering results are the key for an accurate cluster ensemble \citep{Strehl}. Accurate initial partition suggests that the \textit{initKmix} algorithm is able to create accurate and diverse clustering results in different runs. Using each attribute for creating initial clusters in different runs of the KMCMD algorithm could be the reason for it.}

 \subsection{Effect of k on the performance of the KMCMD algorithm with the \textit{initKmix} algorithm} \textcolor{black}{In the experiments (Section 4.1 - Section 4.3) for each  dataset, the number of the desired clusters ($k$) was equal to the number of the actual classes. We also carried out experiments to observe the performance of the KMCMD algorithm with the \textit{initKmix} algorithm when the number of desired clusters was not equal to the number of actual classes. We selected a mixed dataset, Australian credit, for our experiment. The number of the actual classes was two for the dataset. We carried out experiments with $k$ = 4, 5, 6 and 7.  The results are presented in Tables \ref{AustraliancreditAC} - \ref{AustraliancreditARI}. For $AC$ performance measures, we did not observe huge variation as the difference between maximum $AC$ (0.858 for $k$ = 2) and minimum $AC$ (0.823, for $k$ = 6) is 0.035, which is around 4\% of the maximum value (as shown in Table \ref{AustraliancreditAC}). The results demonstrate that the performance of KMCMD with the \textit{initKmix} algorithm is robust to the value of $k$.}
 
 \textcolor{black}{However, the differences between maximum and minimum values are quite large for $RI$ measure (21.56\% of the maximum value) and $ARI$ measure (57.18\% of the maximum value). That was expected as the number of the clusters increases, the data  points in a original class tend to be in different clusters, which leads to lower values of these performance measures. }
 
  \textcolor{black}{We also compared the \textit{initKmix} algorithm against the random initial partition for different values of $k$. Experimental set up was the same as discussed at the start of Section 4. The average and standard deviation of various performance measures are presented in Tables \ref{AustraliancreditAC} - \ref{AustraliancreditARI}. Expect in one case ($k$ = 6 and $AC$ performance measure), KMCMD with the \textit{initKmix} algorithm outperformed KMCMD with the random initial partition in all the cases. }
 
 \textcolor{black}{ The results show that \textit{initKmix} initialization is effective even if $k$ is not equal to the actual number of clusters.}
 
 \begin{table}[!ht]
\centering
\caption{\textcolor{black}{The clustering results ($AC$) of the KMCMD algorithm with \textit{initKmix} and the KMCMD algorithm with the random initial partition method for Australian credit datasets for different numbers of $k$. The better clustering result is shown in bold.}}
 % \begin{tabular}{|p{20mm}|p{20mm}|p{20mm}|p{20mm}|p{20mm}|}
 \begin{tabularx}{\linewidth}{|X|X|X|X|X|}
    \hline
    \multirow{2}{*}{$k$} & \multicolumn{2}{>{\centering\setlength\hsize{2\hsize} }X|}{The KMCMD algorithm  with \textit{initKmix}} & \multicolumn{2}{>{\centering\setlength\hsize{2\hsize} }X|}{The KMCMD algorithm with the random initial partition} \\ 
    %      \multicolumn{2}{c|}{The KMCMD algorithm  with \textit{initKmix}} & 
%      \multicolumn{2}{c|}{The KMCMD algorithm with the random initial partition} \\
   & $\overline{AC}$ & $SD_{AC}$ & $\overline{AC}$ & $SD_{AC}$ \\
    \cline{2-5}%\hline
    2 & \bf{0.858} & 0 & 0.829& 0.118 \\
    \hline
    4 & \bf{0.849} & 0& 0.837 & 0.010 \\
    \hline
    5& \bf{0.852} & 0 & 0.833 & 0.009 \\
    \hline
    6& 0.823 & 0 & \bf{0.835} & 0.011  \\
    \hline
    7 & \bf{0.839} & 0 &0.832  & 0.025 \\
    \hline
  \end{tabularx}
  \label{AustraliancreditAC}
\end{table}
 
 \begin{table}[!ht]
\centering
\caption{\textcolor{black}{The clustering results ($RI$) of the KMCMD algorithm with \textit{initKmix} and the KMCMD algorithm with the random initial partition method for Australian credit datasets for different numbers of $k$. The better clustering result is shown in bold.}}
 % \begin{tabular}{|p{20mm}|p{20mm}|p{20mm}|p{20mm}|p{20mm}|}
 \begin{tabularx}{\linewidth}{|X|X|X|X|X|}
    \hline
    \multirow{2}{*}{$k$} & \multicolumn{2}{>{\centering\setlength\hsize{2\hsize} }X|}{The KMCMD algorithm  with \textit{initKmix}} & \multicolumn{2}{>{\centering\setlength\hsize{2\hsize} }X|}{The KMCMD algorithm with the random initial partition} \\ 
    %      \multicolumn{2}{c|}{The KMCMD algorithm  with \textit{initKmix}} & 
%      \multicolumn{2}{c|}{The KMCMD algorithm with the random initial partition} \\
  & $\overline{RI}$ & $SD_{RI}$ & $\overline{RI}$ & $SD_{RI}$ \\
    \cline{2-5}%\hline
    2 & \bf{0.756} & 0 &0.716 & 0.127 \\
    \hline
    4 & \bf{0.682} & 0& 0.651 & 0.007 \\
    \hline
    5& \bf{0.685} & 0 & 0.648 & 0.019 \\
    \hline
    6 & \bf{0.616} & 0& 0.612 & 0.022  \\
    \hline
    7 & \bf{0.593} & 0 & 0.584 & 0.025 \\
    \hline
  \end{tabularx}
  \label{AustraliancreditRI}
\end{table}

 \begin{table}[!ht]
\centering
\caption{\textcolor{black}{The clustering results ($ARI$) of the KMCMD algorithm with \textit{initKmix} and the KMCMD algorithm with the random initial partition method for Australian credit datasets for different numbers of $k$. The better clustering result is shown in bold.}}
 % \begin{tabular}{|p{20mm}|p{20mm}|p{20mm}|p{20mm}|p{20mm}|}
 \begin{tabularx}{\linewidth}{|X|X|X|X|X|}
    \hline
    \multirow{2}{*}{$k$} & \multicolumn{2}{>{\centering\setlength\hsize{2\hsize} }X|}{The KMCMD algorithm  with \textit{initKmix}} & \multicolumn{2}{>{\centering\setlength\hsize{2\hsize} }X|}{The KMCMD algorithm with the random initial partition} \\ 
    %      \multicolumn{2}{c|}{The KMCMD algorithm  with \textit{initKmix}} & 
%      \multicolumn{2}{c|}{The KMCMD algorithm with the random initial partition} \\
%    & $\overline{AC}$ & SD & $\overline{AC}$ & SD \\
    \cline{2-5}%\hline
    2 & \bf{0.512} & 0 &0.432 & 0.217 \\
    \hline
    4 & \bf{0.368} & 0& 0.307  & 0.014 \\
    \hline
    5& \bf{0.372} & 0 & 0.306 & 0.038 \\
    \hline
    6 & \bf{0.237} & 0&0.228   & 0.044 \\
    \hline
    7 & \bf{0.216} & 0 & 0.191 & 0.049 \\
    \hline
  \end{tabularx}
  \label{AustraliancreditARI}
\end{table}
 
 \subsection{Effect of \textit{n} and \textit{k} on the running time} 
\begin{figure}
  \centering
  \subfigure[Running time for KMCMD algorithm (all the runs)]{\label{runningtime1}\includegraphics[width=0.65\textwidth]{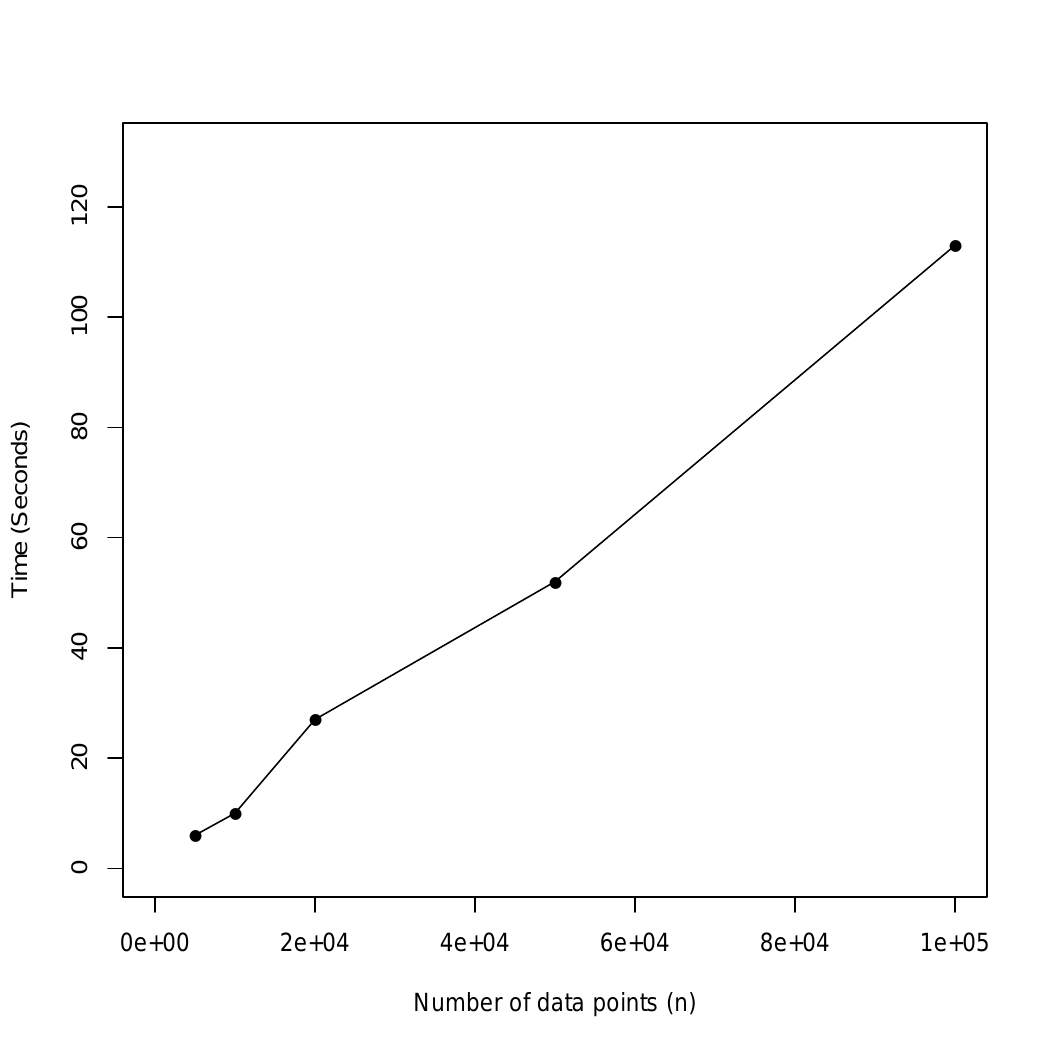}} 
  \subfigure[Running time for combining clustering results ]{\label{numberoctave}\includegraphics[width=0.65\textwidth]{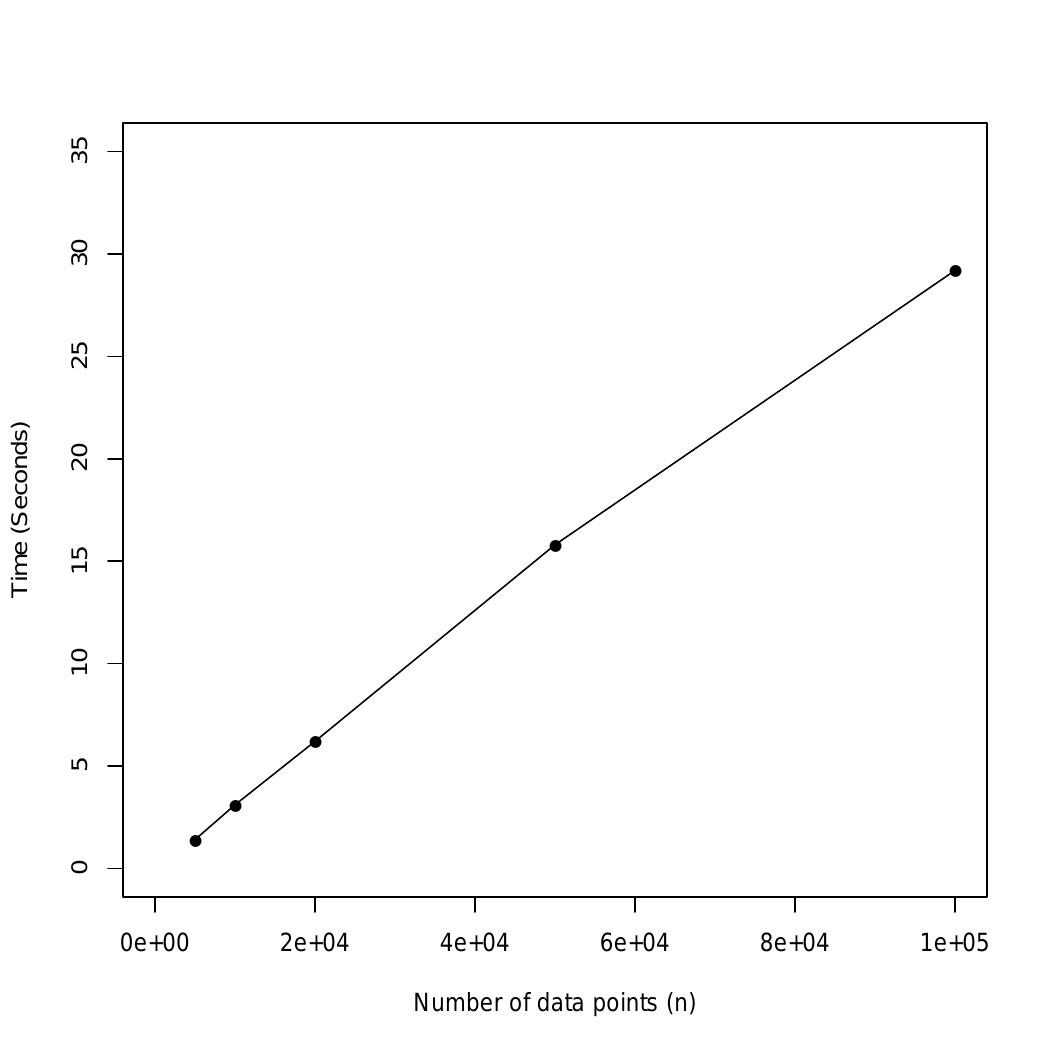}}
  \caption{Running time  vs the number of data points  graphs. The (a) figure presents the time taken by KMCMD algorithm. The (b) figure shows the time taken to combine the clustering results.}
  \label{Combinedfiguresnumbers}
\end{figure}

\textcolor{black}{ We carried out experiments with different datasets to study the effect of \textit{n} and \textit{k} on the running time of the KMCMD algorithm with the \textit{initKmix} algorithm.
 The experiments  were done on a computer with Intel Core i7
1.80 GHz and  16 GB RAM.
To study the effect of $n$ on the running time, 5 artificial datasets of different sizes (5000, 10000, 20000, 50000 and 100000) were generated (by \textit{clustMixType} package \citep{Rclusmix}) using the same procedure as discussed in Section 4.1. These datasets had four attributes and the value of $k$ was set to four. As there were two different parts of the  \textit{initKmix} algorithm (different runs of KMCMD algorithm and cluster ensemble algorithms) and their implementation is in different programming platforms, we present their results in different figures for better understanding. The results presented in Figure \ref{runningtime1} and Figure \ref{numberoctave} demonstrate that the running times of both parts of the \textit{initKmix} algorithm are linear with respect to $n$.}

\begin{figure}
  \centering
  \subfigure[Running time for KMCMD algorithm (all the runs)]{\label{numberkmcmd1}\includegraphics[width=0.65\textwidth]{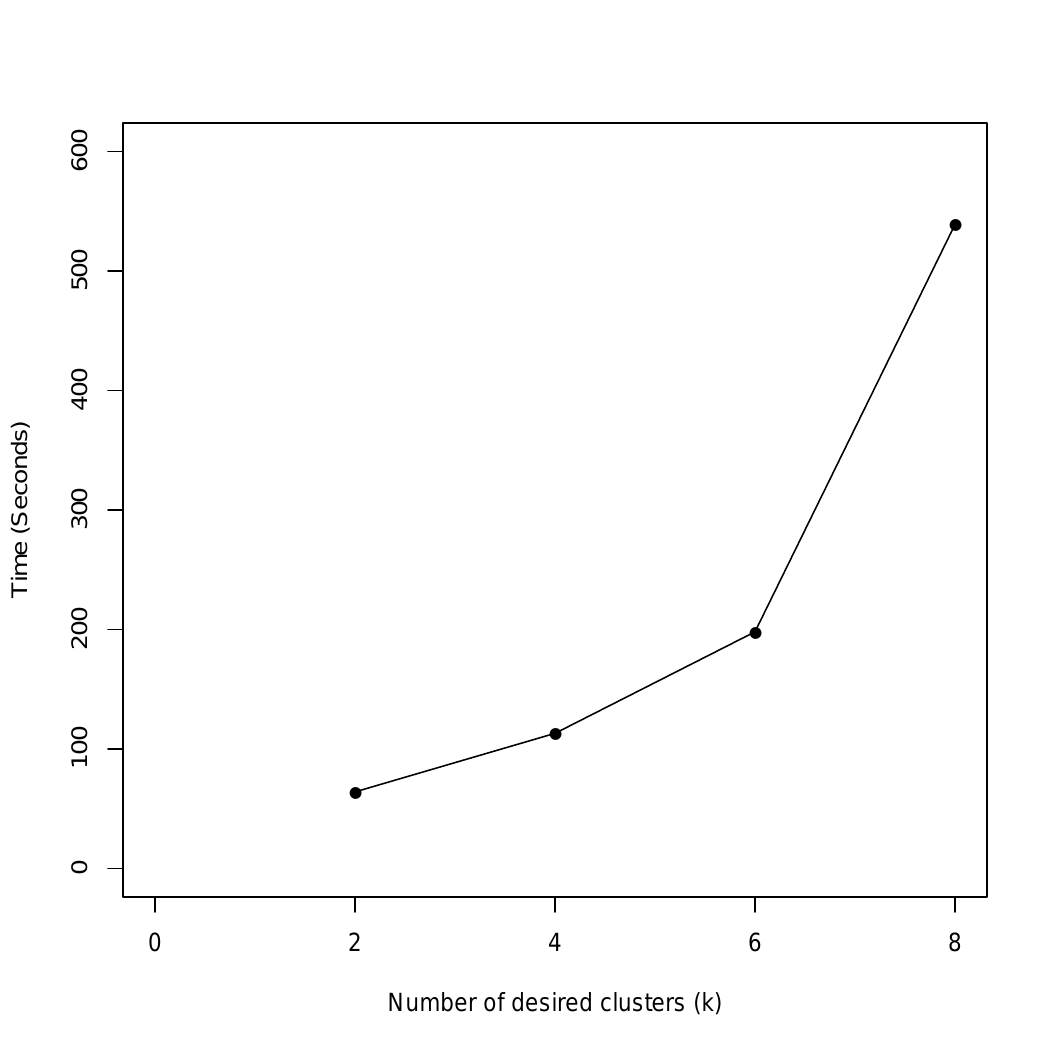}} 
  \subfigure[Running time for combining clustering results ]{\label{numberoctave1}\includegraphics[width=0.65\textwidth]{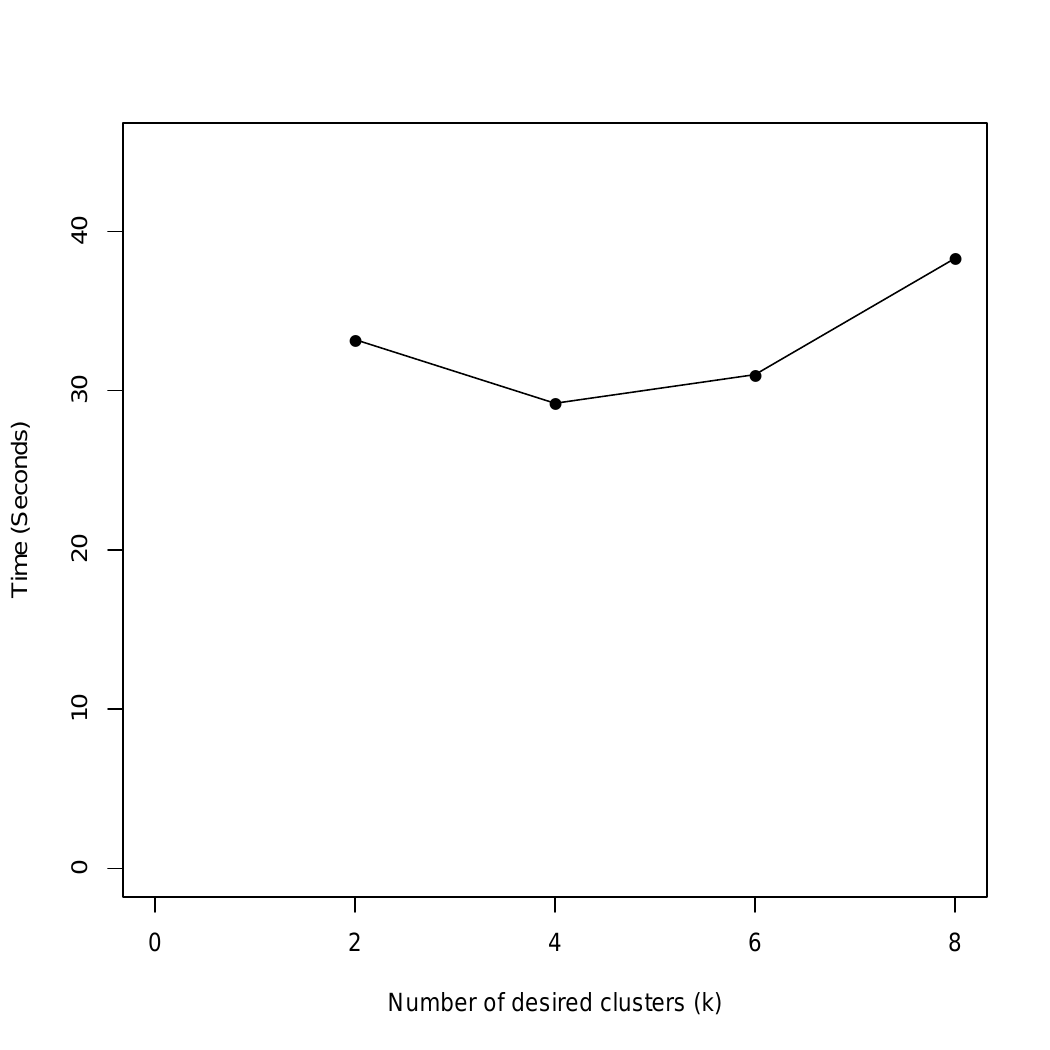}}
  \caption{Running time vs the number of clusters ($k$)  graphs. The (a) figure presents the time taken by KMCMD algorithm. The (b) figure shows the time taken to combine the clustering results.}
  \label{Combinedfiguresclusters}
\end{figure}

\textcolor{black}{We also carried experiments to study the effect of the value of $k$ on the running time of \textit{initKmix} algorithm. We selected the dataset with 100,000 data points for the experiment. Four different values of $k$ (2, 4, 6 and 8) were used in the experiment. The running times for KMCMD algorithm (for all runs) and cluster ensemble algorithm are presented in Figure \ref{numberkmcmd1} and Figure \ref{numberoctave1}. The results suggest that there is a increase in the running time of KMCMD algorithm; however, no significant changes were observed for cluster ensemble algorithms. We can infer from these running times that the total time of \textit{initKmix} algorithm increases with the value of $k$. It is to be noted that we run KMCMD algorithm $m$ time using each attribute as initial clusters, the value of desired clusters in a run with a categorical attribute is set to the number of unique attribute values. Therefore, it is independent of the value of $k$. In the present case, two out of four instances of clustering results are the same for all the values of $k$.}

\subsection{Analysis of individual clustering results} 
Regarding the KMCMD algorithm with the \textit{initKmix} algorithm, we run the KMCMD algorithm $m$ times to produce $m$ clustering results. These $m$ results are combined to yield the initial partition and then the KMCMD algorithm is run with this initial partition to obtain the final clustering results. Therefore, we perform an analysis to compare the accuracy of $m$ individual clustering results and the final clustering result using the $AC$ performance measure.

For this analysis, we selected two categorical datasets, Vote and Mushroom, and two mixed datasets, Heart (Statlog) and Australian credit, for this analysis. For better comparative study, for categorical datasets, we selected these datasets with some attributes having the same number of values as the number of desired clusters. Similarly, for mixed datasets, we selected these datasets with some categorical attributes having the same number of values as the number of desired clusters. Using categorical attributes, with the numbers of attribute values not equal to the desired clusters, to create the initial clusters does not produce the desired number of clusters. Therefore, clustering results when those attributes were used as initial clusters were not selected for the comparative study.

\begin{figure}[!htb]
  \centering
  \subfigure[Vote]{\label{votefig}\includegraphics[width=0.4\textwidth]{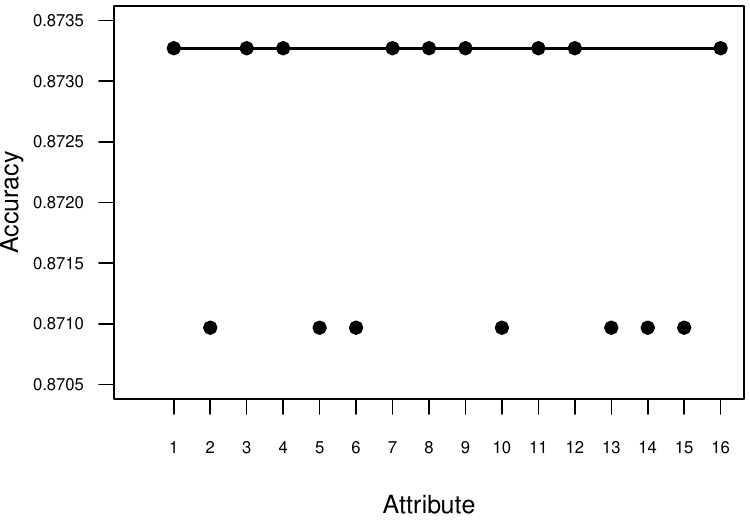}} 
  \subfigure[Mushroom]{\label{Mushroomfig}\includegraphics[width=0.4\textwidth]{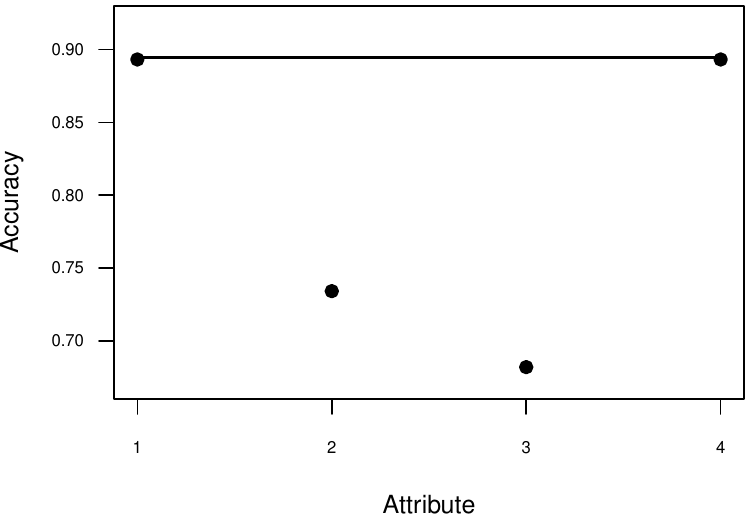}}
  \subfigure[Heart (statlog)]{\label{Heartfig}\includegraphics[width=0.4\textwidth]{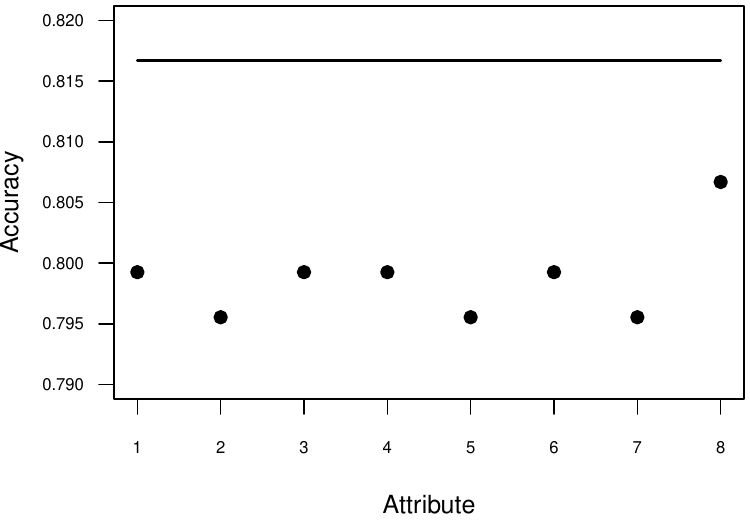}}
 \subfigure[Australian Credit]{\label{Auscreditfig}\includegraphics[width=0.4\textwidth]{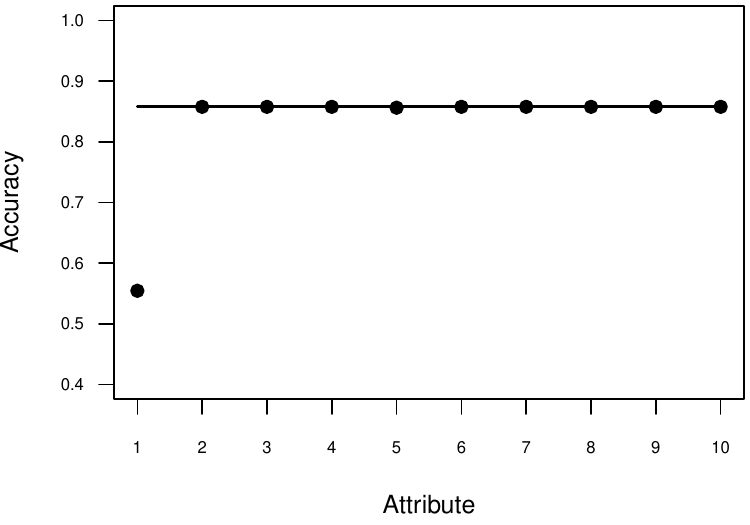}} 
  \caption{Clustering results for various datasets. A dark filled circular dot shows the performance for the KMCMD algorithms by using that attribute for obtaining initial partition. The straight line represents the final clustering result obtained with the \textit{initKmix} algorithm. The dark filled circular dot below the line shows that the performance in that run is worse than the final result. The dark filled circular dot on the line shows that the performance in that run is similar to the final result.}
  \label{Combinedfigures}
\end{figure}

For Vote dataset, the individual clustering results for $16$ categorical attributes as initial clusters in different runs are presented in  Figure \ref{votefig}. We did not observe large differences in individual clustering results (the minimum $AC$ - $0.8709$ and the maximum $AC$ - $0.8732$). The final $AC$ ($0.8732$) was equal to the maximum $AC$. For the Mushroom dataset, only four attributes had two values (same as the number of desired clusters). The results for these four attributes are presented in  Figure \ref{Mushroomfig}. There was a large variation in individual clustering results (the minimum $AC$ - $0.682$
, the maximum $AC$ - $0.893$). The final clustering result ($0.894$) was slightly better than the best individual clustering ($0.893$).
Figure \ref{Heartfig} has individual clustering results for the {\it{Heart (Statlog)}} dataset (the minimum $AC$ - $0.795$, the maximum $AC$ - $0.807$), the final clustering accuracy was $0.817$, which was slightly better than the best individual accuracy (0.807). The individual clustering results for {\it{Australia credit}} dataset (for six numeric and four categorical) are presented in Figure \ref{Auscreditfig} (the minimum $AC$ - $0.554$
, the maximum $AC$ - $0.858$). There was a large variation in individual clustering results. The final clustering result ($0.858$) was equal  to the best individual clustering result.

The analysis suggests that individual clustering results for various datasets had small or large variations; however, the final clustering results were equal to or better than the best individual clustering results. This shows that combining clustering results is a good approach to obtain better initial clusters, and when feeding these  to the KMCMD algorithm results in better clustering accuracy for the studied categorical and mixed datasets.

\section{Conclusion and Future Work} 
\label{sec:conclusions}
KMD algorithms suffer from the random initial partition problem that can lead to different clustering results in different runs thereby undermining the reliability of results. In this paper, we presented \textit{initKmix,} an algorithm to find the initial partition for the KMCMD algorithm \citep{AmirLipika2007}. The algorithm uses an individual attribute to create the initial partition when running the KMCMD algorithm. Multiple clustering results created by this procedure are combined to obtain the initial partition.  The clustering results obtained using the KMCMD algorithm  with the initial partition created by \textit{initKmix} were accurate and consistent. The KMCMD algorithm with \textit{initKmix}  algorithm outperformed KMCMD algorithm with the random initial partition method  on multiple categorical and mixed datasets. The KMCMD algorithm with \textit{initKmix}  algorithm also performed similar to or better than other state-of-the-art clustering algorithms on multiple categorical and mixed datasets. \textcolor{black}{Computational complexity analysis and running time results suggest that the running time of the KMCMD algorithm with \textit{initKmix}  is linear with respect to number of data points. Therefore, this clustering algorithm scales well with the number of data points. Results also demonstrated that the performance of the KMCMD algorithm with the \textit{initKmix} algorithm is robust to the choice of the $k$.}

In future, other KMD algorithms \citep{Huang1997,AutomatedWeightclus,Modhaclustering} with  \textit{initKmix} will also be studied.
 KMD algorithms have been suggested for fuzzy clustering \citep{ji2012fuzzy,DU201746} and subspace clustering \citep{Ahmadsubspace}, in future, we will consider applying the \textit{initKmix} algorithm to these clustering algorithms. We will also investigate possible extension of \textit{initKmix} to find the value of \textit{k} for KMD algorithms.

\section*{Funding}{This research was funded  by a UAE university Start-up grant (grant number G00002668; fund number 31T101).}
\section*{Conflicts of interest}{The authors declare no conflict of interest.} 
%\reviewreports{\\
%Reviewer 1 comments and authors’ response\\
%Reviewer 2 comments and authors’ response\\
%Reviewer 3 comments and authors’ response
%}
%%%%%%%%%%%%%%%%%%%%%%%%%%%%%%%%%%%%%%%%%%
\bibliography{mixeddataref}
\end{document}